\documentclass[conference]{IEEEtran}
\IEEEoverridecommandlockouts

\usepackage{cite}
\usepackage{amsmath,amssymb,amsfonts}
\usepackage{algorithmic}
\usepackage{graphicx}
\usepackage{graphics}
\usepackage{multirow}
\usepackage{textcomp}
\usepackage{xcolor}
\usepackage{authblk}
\usepackage{subfigure}
\usepackage{tabularx}
\usepackage{booktabs}

\def\BibTeX{{\rm B\kern-.05em{\sc i\kern-.025em b}\kern-.08em
    T\kern-.1667em\lower.7ex\hbox{E}\kern-.125emX}}

\newcommand{\paperid}{1076}

\makeatletter
\newif\if@blind

\@blindfalse  

\title{Hinting Pipeline and Multivariate Regression CNN for Maize Kernel Counting on the Ear
    \if@blind

    \else
    \thanks{This work was performed as part of the \textit{PlatIAgro} project, which is conducted by CPQD in partnership with RNP (National Teaching and Research Network), and funded by the Ministry of Science, Technology and Innovation of Brazil.}\fi}

\if@blind
    \author{Blind submission - Paper ID \#\paperid}
\else
\author[1]{Felipe Araújo}
\author[1]{Igor Gadelha}
\author[2]{Rodrigo Tsukahara}
\author[1]{Luiz Pita}
\author[1]{Filipe Costa}
\author[1]{Igor Vaz}
\author[1]{\\Andreza Santos}
\author[1]{Guilherme Folego}

\affil[1]{CPQD - Artificial Intelligence and IoT Solutions, Brazil}
\affil[2]{Fundação ABC, Brazil}

\fi
\makeatother

\begin{document}
\begin{sloppypar}
\hyphenpenalty=10000

\maketitle

\begin{abstract}
Maize is a highly nutritional cereal, widely used for human and animal consumption and also as raw material by the biofuels industries. This highlights the importance of precisely quantifying the corn grain productivity in season, helping the commercialization process, operationalization, and critical decision-making. Considering the manual labor cost of counting maize kernels, we propose in this work a novel preprocessing pipeline named \textit{hinting} that guides the attention of the model to the center of the corn kernels and enables a deep learning model to deliver better performance, given a picture of one side of the corn ear. Also, we propose a multivariate CNN regressor that outperforms single regression results. Experiments indicated that the proposed approach excels the current manual estimates, obtaining $MAE$ of $34.4$ and $R^2$ of $0.74$ against $35.38$ and $0.72$ for the manual estimate, respectively.

\end{abstract}

\begin{IEEEkeywords}
Corn kernel counting, Hinting pipeline, CNN, Multivariate regression model
\end{IEEEkeywords}

\section{Introduction}
\label{sec:intro}

Maize (\textit{Zea mays L.}) is considered one of the most important cereals in the world due to its high production potential, chemical composition, and nutritional value. These qualities result in wide use for animal and human food, and drive high-tech industries to produce biofuels~\cite{corn_intro}.

Currently, the process of quantifying maize grain production per unit of area in field conditions is laborious and time-consuming. The development of fast methods that are equivalent in accuracy and precision, based on computer vision algorithms, has already proved effective in some studies~\cite{khaki2020convolutional,li2019corn}. They can significantly improve the sampling process in time and space, also considering the most heterogeneous crops, and the best estimation of the volume of mass and quantity of grains within a given plot. 

\begin{figure}[t]
    \begin{center}
        \includegraphics[width=0.9\columnwidth]{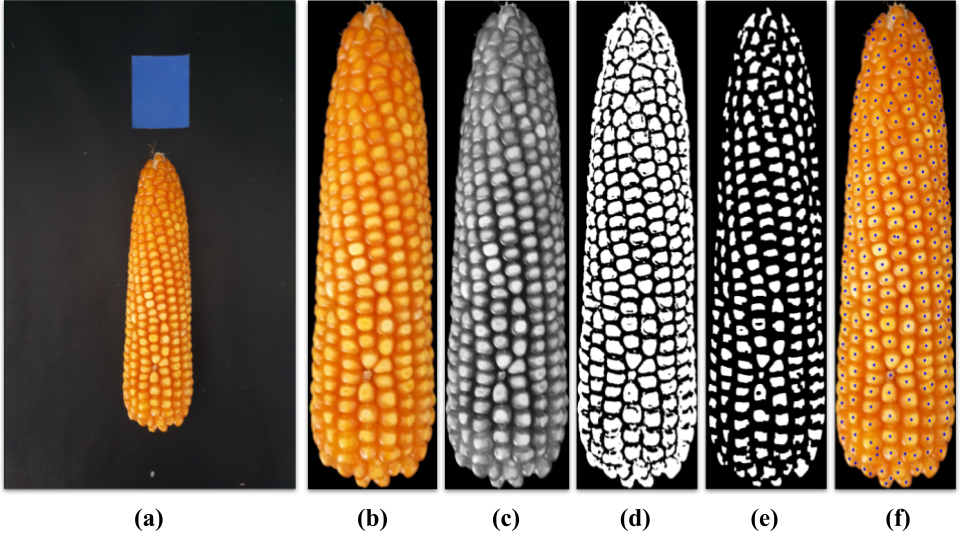}
        \caption{Hinting Pipeline steps: (a) original image; (b) maize kernel segmentation; (c) grayscale conversion and image improvement; (d) thresholding; (e) morphologic operations; (f) maize kernel center marking taking the connected components centroids. The blue Post-it on the figure was not used in this work.}
    \label{fig:hints-pip}
    \end{center}
\end{figure}

In this sense, in this work, we propose the design and development of a novel preprocessing pipeline named \emph{hinting}, depicted in Figure~\ref{fig:hints-pip}, which guides the attention of the model to the center of the maize kernels and enables a deep learning model to deliver better performance, assuming only pictures of a single side of each ear, which is an important characteristic. We also designed a Multivariate Convolutional Neural Network (CNN) regression model for quantifying kernels of maize.

For this task, we created a dataset with pictures of a single side of each ear, considering one ear per image, cultivated in two contrasting edaphoclimatic regions, considering a wide variability of genetic materials used in the South and Southeast regions of Brazil.

Thus, the main contributions of this work are:

\begin{itemize}
  \item a dataset with pictures of a wide variability of genetic materials used in the South and Southeast regions of Brazil;
  \item a color- and contour-based image preprocessing pipeline to extract background and generate a maize image dataset;
  \item a multivariate CNN regressor for maize kernel counting on the ear, assuming only one single side of each corn ear;
  \item an image preprocessing pipeline to improve deep learning results,
  named \textit{hinting};
  \item the automation of a laborious and time-consuming job achieving better results with less time and effort.
\end{itemize}

This article is organized as follows. Section~\ref{sec:related_works} presents related works for maize kernel counting. The proposed approach is detailed in Section~\ref{sec:proposed_approach}. Experiments and their results are described in Sections~\ref{sec:experiments} and~\ref{sec:results-discussion}, respectively. Finally, we conclude the paper in Section~\ref{sec:conc}.

\section{Related Works}
\label{sec:related_works}

Previous works adopting digital image analysis and deep learning for maize kernel counting were based on images of maize kernels on plain surfaces, already removed from the ears~\cite{severini2011cropscience, zhang2011tscae, khaki2020convolutional}.

Miller et al.~\cite{miller2017plantjournal} retrieved information about the average space of kernels along the cob axis, the number of kernels, and measures about the kernel's size axis. 
Li et al.~\cite{li2019corn} proposed a maximum likelihood estimator algorithm to classify normal and damaged maize kernels using images of kernels removed from the cob, handcrafted features, and line profile-based segmentation (LPSA) to isolate touching maize kernels. 
Grift et al.~\cite{grift2017biosystems} proposed a semi-automated approach to estimate the kernels in a corn ear. The authors built a special box to photograph the ears using controlled lighting, background, and camera conditions. This box can also rotate the ear to place all kernels in front of the camera. The approach segments kernels by applying Otsu thresholding, morphological operations, kernel's pixel area, and kernel's center of mass. 
Khaki et al.~\cite{khaki2020convolutional} used a Deep Learning approach for detecting and counting maize kernels in a single corn ear image. First, a detection algorithm is performed to find the regions with maize kernels. Then, the authors use a regression CNN model to determine $(x, y)$ coordinates of the center of the kernels from the windows classified as positive by the previous step. Finally, they count the number of kernel centers found on the corn image. An improvement was reported on~\cite{khaki2021deepcorn}, in which the authors calculated density maps and image threshold processing and used this output to count the number of kernels in a corn ear image. To estimate the number of kernels in the ear, the authors multiply the kernel counting by a constant.

Different from the aforementioned works, we propose an approach to estimate the total number of kernels of maize in a semi-controlled environment using only a one-sided picture of the maize. We require a dark background picture with a standard digital camera or smartphone.
The estimation is performed using a multivariate CNN regressor, which will be presented in the next section.

\section{Proposed Approach}
\label{sec:proposed_approach}

This work proposes a new approach for estimating the total number of maize kernels based on a CNN regression model. To fully evaluate the proposed approach, we conducted several experiments to analyze the ability of the method to predict the number of on-ear corn kernels based on a single-side input image.
\subsection{Hinting Pipeline}
\label{subsec:hinting}

The Hinting Pipeline consists of a set of processing image operations that indicates the center of every kernel of maize. This pipeline performs according to the following steps, depicted in Figure~\ref{fig:hints-pip}.

\begin{itemize}
\item \textbf{Maize kernel segmentation:} We segment the regions with pixels representing the yellow color, defined by a range in the Hue channel of HSV space. Then, we create a binary mask by selecting the largest connected component to extract the ear from the original image.
\item \textbf{Grayscale and image improvement:} we convert the input image to grayscale and improve its brightness and contrast using Contrast Limited Adaptive Histogram Equalization (CLAHE)~\cite{10.5555/992362.2845017}. We also remove noise by applying median filtering;

\item \textbf{Thresholding:} We binarize the image through an adaptive threshold and perform a bitwise AND operation between the thresholded image and the binary mask to remove the background.

\item \textbf{Morphological operations:} We apply morphological operations over the generated binary image to remove some artifacts caused by the previous process and to evidence the contours of the kernels in the maize ear. 

\item \textbf{Maize kernel center marking:} Finally, we detect the center coordinates of all contours on the ear in the resulting image of the pipeline and mark these coordinates in the image.

\end{itemize}

\subsection{Multivariate CNN Regressor}
\label{cnn}
We developed a CNN model that receives the output of the hinting pipeline for an image of one side of the maize ear and outputs an estimate of the total number of kernels. To this end, we implemented a custom residual CNN architecture~\cite{resnet} to perform a regression task.

\begin{figure}[t]
  \centering
  \includegraphics[width=0.9\columnwidth]{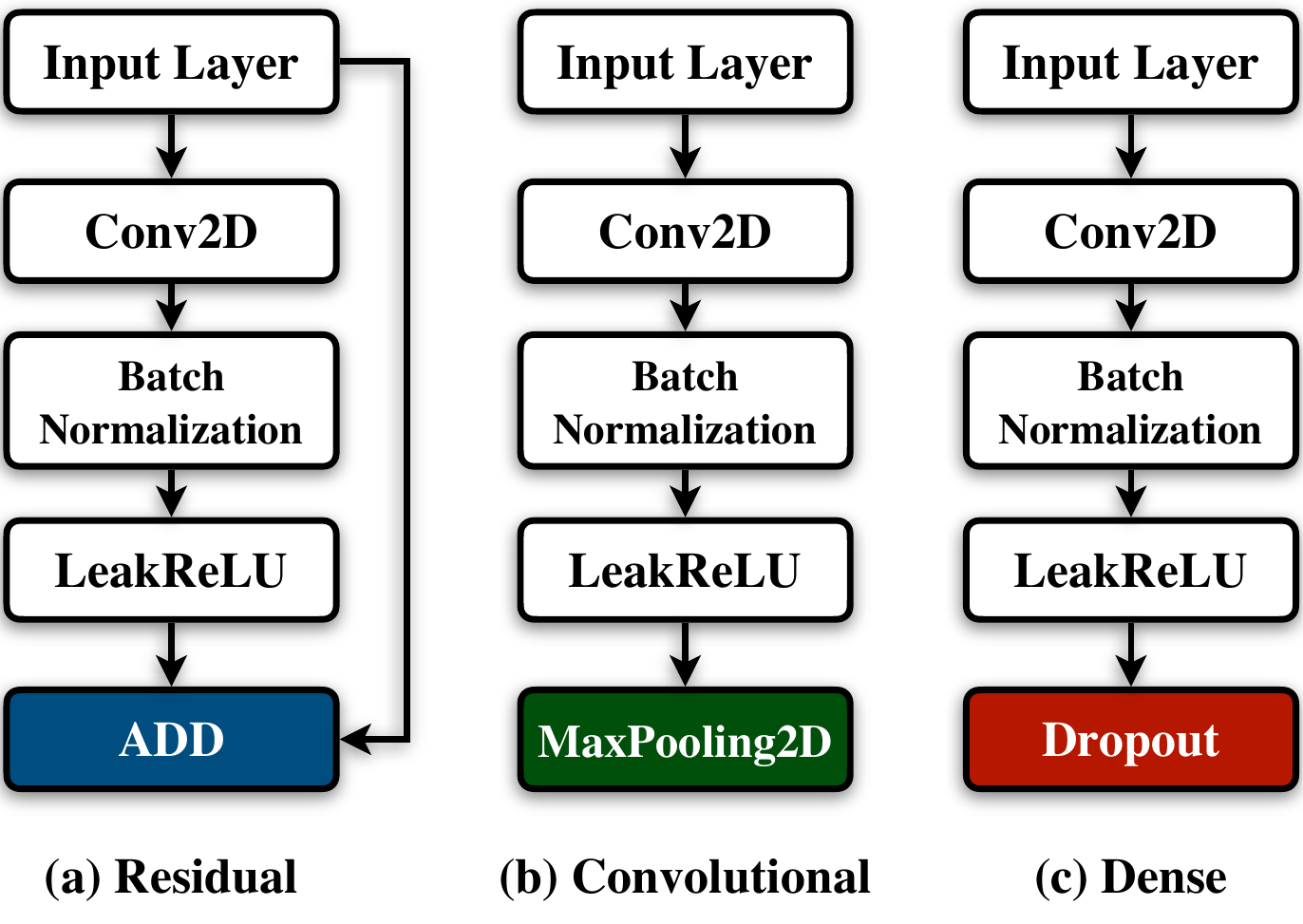}
\caption{Implemented blocks for custom residual architecture}
\label{fig:cnnblock}
\end{figure}

For this, we implemented the residual block as depicted in Figure~\ref{fig:cnnblock}a. First, a convolutional layer extract features from the input layer. Then, the extracted features are normalized with batch normalization and processed by the Leaky ReLU non-linear activation function. Finally, the output of the activation function is summed with the input of the residual block, creating an effective residual connection that allows the gradient flow throughout the residual block.

Besides the residual blocks, we also used standard convolutional blocks composed of a convolutional layer followed by batch normalization, Leaky ReLU, and Max Pooling, as depicted in Figure~\ref{fig:cnnblock}b. 

The convolutional and residual blocks are concatenated to constitute a \emph{combined} block. We configured our residual architecture with six combined blocks of $32$, $64$, $128$, $256$, $512$, and $1024$ channels. We also applied a global average pooling in the output of the last combined block.

Finally, a dense block (Figure~\ref{fig:cnnblock}c) formed by a dense layer followed by batch normalization, Leaky ReLU (factor of $0.3$), and Dropout (factor of $0.2$), processes the features, and another dense layer with linear activation provides the regression output of the model.

\begin{table}[htb]
\centering
\caption{Results summary for the total number of maize kernels estimation output.}
\resizebox{0.5\textwidth}{!}{
\begin{tabular}{l||ll|cccccccc}
\textbf{Experiment} & \textbf{Metric} & \textbf{Group} & \textbf{Mean} & \textbf{Std} & \textbf{Min} & \textbf{Median} & \textbf{Max} \\
\multirow{4}{*}{Multivariate} & \multirow{2}{*}{\textit{R²}} & Baseline & 0.70 & 0.02 & 0.64 & 0.69 & 0.73 \\
 &  & Multivariate & \textbf{0.72} & 0.02 & 0.68 & 0.72 & 0.76\\ \cline{2-8}
 & \multirow{2}{*}{MAE} & Baseline & 38.3 & 1.53 & 35.5 & 35.7 & 41.9  \\
 &  & Multivariate & \textbf{36.0} & 1.63 & 32.6 & 38 & 40.1 \\ \midrule
\multirow{6}{*}{Hinting Pipeline} & \multirow{3}{*}{\textit{R²}} & Multivariate & 0.72 & 0.02 & 0.68 & 0.72 & 0.76 \\
 &  & Control & 0.71 & 0.02 & 0.67 & 0.71 & 0.75 \\
 &  & Hints & 0.74 & 0.02 & 0.68 & 0.74 & 0.77 \\ \cline{2-8}
 & \multirow{3}{*}{MAE} & Multivariate & 36.0 & 1.63 & 32.6 & 35.7 & 40.1\\
 &  & Control & 37.0 & 1.76 & 33.8 & 36.9 & 39.9\\
 &  & Hints & 34.8 & 1.94 & 32.1 & 34.4 & 39.9\\ \midrule
\end{tabular}
}
\label{tab:results}
\end{table}

We also verified the performance of the proposed approach with additional outputs as a multivariate regression.
We used the image also to predict the number of vertical rows in the entire ear and the number of kernels in two vertical rows. Such modification was done on the last dense layer, changing the number of outputs from one to four. The hypothesis is that providing more information to the model will improve the optimization process to boost performance results.

\section{Experimental Setup}
\label{sec:experiments}
This section presents details about the experiments we performed in this work. We detail the proposed dataset, the evaluation metrics, and some model training details. 

\subsection{Image Data Collection}
\label{subsec:dataset}

Two gatherings were carried out during the 2019/2020 growing season.  The first gathering consisted of $46$ commercial maize hybrids in Castro (Paraná, Brazil), $1050m$ altitude, and Cfb Köppen-Geiger climate classification~\cite{alvares2013koppen}. The second gathering contained $17$ commercial hybrids and was carried out in Itaberá (São Paulo, Brazil), $700m$ altitude, and Cfa Köppen-Geiger climate classification.

Five maize ears of each hybrid were randomly collected between the physiological stage of maturation and the harvest date. These samples were placed on a dark frame for photographic capture with a $12$ MP camera embedded in an Android $8.0$ smartphone. 
The maize ears were rotated 180 degrees on their long axis, and a new image was acquired, resulting in two photos per ear.

To estimate the performance of the proposed approach to unseen data, we divided the whole dataset into three distinct groups without overlap. We split the ears following the proportions of $60\%$, $20\%$, and $20\%$ used to train, validate, and test, respectively. We also employed a stratification procedure to guarantee that we have at least one sample of each hybrid in each of the three subsets. In this manner, the training subset comprises $189$ maize ears from $50$ hybrids, leading to $378$ images being front and back of the maize ear. The evaluation and test subsets include $63$ maize ears from the same $50$ hybrids and a total of $126$ images for each group.

\subsection{Evaluation Metrics}

We follow the standard guidelines generally used to evaluate regression models to estimate the behavior of our approach. Therefore, we use the Mean Absolute Error (MAE), and R-squared ($R^2$), to assess the regression quality in the three distinct subsets of the dataset.

\subsection{Training Details}

Each RGB image of the dataset was preprocessed by cutting the images to centralize the maize ear and resizing them to obtain images with size $512 \times 128$. This way, we maintained the average aspect ratio of a maize ear. Vertical and horizontal random flip is also used in a data augmentation strategy to increase the number of diverse images. The data was grouped in batches of size $64$, considering a random shuffling for the training dataset.

We trained each model for $100$ epochs, keeping the model in the epoch that got the best $R^2$ score in the validation set. We used Adam optimizer~\cite{adam}, starting with a learning rate of $0.0001$. We also used a reduce-on-plateau strategy, with a factor of $0.1$, to guarantee a better convergence of the gradient. The training loss used to update the gradient was the Mean Absolute Error (MAE).
The training and testing steps were done using an NVIDIA Tesla T4 GPU with $16$ GB of RAM.

\section{Results and discussion}
\label{sec:results-discussion}

We present the results in three subsections: multivariate regression, hinting pipeline, and manual counting \textit{vs.} CNN regression. To better estimate the global differences between the compared approaches, each CNN model was trained $30$ times. We summarize the obtained results for the total number of kernels estimation output in Table~\ref{tab:results}.

\begin{figure}[t!]
     \begin{center}
         \includegraphics[width=0.9\columnwidth]{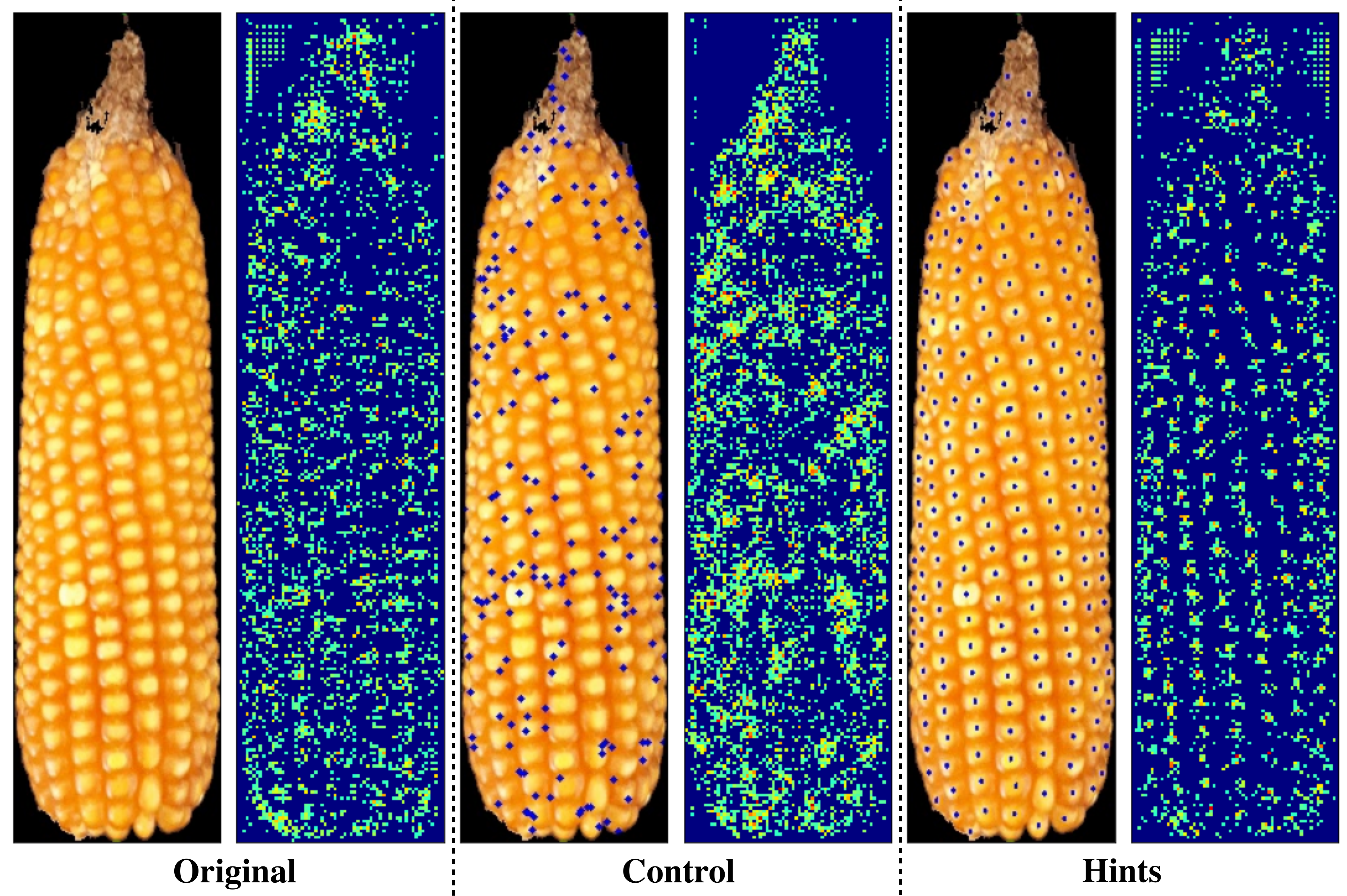}
         \caption{Salience maps of a sample of our database. We evaluated three cases in our study:  Original image, Control (Random Dots), and \textit{Hints} (Section~\ref{subsec:hinting}).}
         \label{fig:saliency-map}
    \end{center}
 \end{figure}

\subsection{Multivariate Regression}

First, we explore the effects of including additional outputs. We refer to the standard approach, with a single image as input and a single regression variable (total number of kernels) as output, as the \textit{baseline} regression. Similarly, we nominate the regressor with multiple outputs as \textit{multivariate}. In this approach, the model receives a single image and approximates multiple outputs, \textit{i.e}, the total number of kernels, the number of rows, and the number of kernels in two  of these rows. Results are presented in Table~\ref{tab:results}.

A Mann-Whitney test indicated a greater $R^2$ score for the proposed regression approach with multiples outputs ($Mdn = 0.72$) than the baseline regression model ($Mdn = 0.69$), $U = 189$, $p < .001$. The multivariate also presented a lower estimated distance from the true MAE value ($Mdn = 35.7$) compared to the \textit{baseline} model ($Mdn = 38$), $U = 130$, $p < .001$. We argue that modeling additional information correlated to the desired output increases the quality of the gradient throughout the network layers, increasing the efficiency of the training and leading to a better approximation. We perform the following experiments with the multivariate model, given its better results.

\subsection{Hinting pipeline}

To validate our hinting pipeline hypothesis, we trained the model using three different preprocessing techniques and applied the proposed multivariate regression format.

We start with a standard preprocessing pipeline as a baseline approach, which is basically the maize kernel segmentation step described in Section~\ref{subsec:hinting}. Then, we create a control group with blue dots added randomly to the baseline image. The number of blue dots added is equal to the median value of the total number of kernels from all images in the training dataset ($240$).
Finally, we employ the full hinting pipeline detailed in Section~\ref{subsec:hinting}.

Results of this experiment are reported in Table~\ref{tab:results}. Kruskal-Wallis H-test~\cite{vargha1998kruskalhtest} indicated a significant difference between the groups for $R^2$ metric, $H = 23.5$, $p < .001$, and also indicated a significant difference between groups for MAE metric, $H = 19$, $p < .001$. Post-hoc analyses using Mann-Whitney test for pairwise group comparison ~\cite{mann1947test} indicated that the $R^2$ metric of the control group ($Mdn = 0.71$) performed worse than the models trained using the standard ($Mdn = 0.72$), $U = 291$, $p < 0.01$ and hint pipeline ($Mdn = 0.74$), $U = 144$, $p < .001$. The same behavior was observed for the MAE metric, where the control group achieved the largest prediction distance ($Mdn = 36.9$) compared to the standard ($Mdn = 35.7$), $U = 321$, $p = .028$, and hinting pipelines ($Mdn = 34.4$), $U = 170$, $p < .001$.
Secondly, the hints played an important role in helping the model better generalize maize kernels' counting.

To assess the relevance of the hinting preprocessing procedure for the model decision-making, we perform a qualitative analysis of the corn images from our train set, calculating their salience maps~\cite{simonyan2013deep} and evaluating them. Figure~\ref{fig:saliency-map} depicts one sample of corn and its salience map, assuming the aforementioned preprocessing scenarios. This visual investigation reinforces our expectations that the hinting procedure improved the network attention for the pixels located around the center of kernels, allowing us to achieve better results, increasing the performance of the recognition of corn kernels and contributing to a higher generalization within unseen data represented by the test set.

\subsection{Manual counting \textit{vs.} regression CNN}

The manual estimation of the total number of kernels uses an approximation rule based on the number of kernels in two rows and the number of rows in a maize ear, given by Equation~\ref{eq:manual_count}.

\begin{equation}
    \label{eq:manual_count}
    \tiny
    \textrm{TOTAL KERNELS} =  \frac{\textrm{KERNELS IN ROW $1$} + \textrm{KERNELS IN ROW $2$}}{2} \times \textrm{NUMBER OF ROWS} 
    \normalsize
\end{equation}

Table~\ref{table:manualvscnn} presents a comparison between the proposed approach and the manual approximation obtained by applying the above equation to the test dataset.

\begin{table}[t!]
\centering
\renewcommand{\arraystretch}{1.3}
\setlength{\tabcolsep}{2.5pt}
\caption{Comparison between CNN Regressor and Manual Approximation}
\resizebox{0.8\columnwidth}{!}{\begin{tabular}{c|c|c}
\hline
\textbf{Metric} & \textbf{Manual Approximation} & \textbf{CNN Regressor (Mdn)} \\ \hline
$R^2$              & $0.72$                 & $0.74$                  \\
MAE & $35.38$ & $34.40$ \\

\end{tabular}}
\label{table:manualvscnn}
\end{table}

This means that our method is not only faster, requiring less manual labor, but it also gets closer to the real count value compared to the manual approximation.

\section{Conclusions}
\label{sec:conc}

In this paper, we investigated the use of a multivariate CNN regression model for quantifying kernels of maize, considering a wide variability of genetic materials. The proposed Hinting Pipeline combined with the designed CNN-based approach has demonstrated the potential to solve plant phenotyping issues, in particular counting kernels in maize ears. Results from experiments reveal the feasibility of determining the number of maize kernels per ear from an image taken with a mobile phone.

To the best of our knowledge, this is the first work that enhances the performance of a multivariate regression model through a hinting pipeline technique for maize kernel counting. Our findings showed that the hinting pipeline guided the attention of the model for the center of the ear kernels and led the model to achieve significant results related to the computed metrics.

Finally, we highlight the need to conduct new studies focusing on the thousand-grain weight of maize. This variable, along with the count of kernels per ear is the most important in the sensitivity analysis of the corn yield equation.

\section*{Acknowledgments}
\label{sec:ack}

This work was performed as part of the \textit{PlatIAgro} project, which is conducted by CPQD in partnership with RNP (National Teaching and Research Network) and funded by the Ministry of Science, Technology and Innovation of Brazil.

\newpage

\bibliographystyle{IEEEtran}
\bibliography{main.bib}

\begin{thebibliography}{10}
\providecommand{\url}[1]{#1}
\csname url@samestyle\endcsname
\providecommand{\newblock}{\relax}
\providecommand{\bibinfo}[2]{#2}
\providecommand{\BIBentrySTDinterwordspacing}{\spaceskip=0pt\relax}
\providecommand{\BIBentryALTinterwordstretchfactor}{4}
\providecommand{\BIBentryALTinterwordspacing}{\spaceskip=\fontdimen2\font plus
\BIBentryALTinterwordstretchfactor\fontdimen3\font minus
  \fontdimen4\font\relax}
\providecommand{\BIBforeignlanguage}[2]{{%
\expandafter\ifx\csname l@#1\endcsname\relax
\typeout{** WARNING: IEEEtran.bst: No hyphenation pattern has been}%
\typeout{** loaded for the language `#1'. Using the pattern for}%
\typeout{** the default language instead.}%
\else
\language=\csname l@#1\endcsname
\fi
#2}}
\providecommand{\BIBdecl}{\relax}
\BIBdecl

\bibitem{corn_intro}
T.~Shah, K.~Prasad, and P.~Kumar, ``Maize -- a potential source of human
  nutrition and health: A review,'' \emph{Cogent-Food and Agriculture}, vol.~2,
  pp. 1--9, 2016.

\bibitem{khaki2020convolutional}
S.~Khaki, H.~Pham, Y.~Han, A.~Kuhl, W.~Kent, and L.~Wang, ``Convolutional
  neural networks for image-based corn kernel detection and counting,''
  \emph{MDPI Sensors}, vol.~20, no.~9, p. 2721, 2020.

\bibitem{li2019corn}
X.~Li, B.~Dai, H.~Sun, and W.~Li, ``Corn classification system based on
  computer vision,'' \emph{MDPI Symmetry}, vol.~11, no.~4, p. 591, 2019.

\bibitem{severini2011cropscience}
A.~D. Severini, L.~Borrás, and A.~G. Cirilo, ``Counting maize kernels through
  digital image analysis,'' \emph{Crop Science}, vol.~51, no.~6, pp.
  2796--2800, 2011.

\bibitem{zhang2011tscae}
Y.~Zhang, W.~Wu, and G.~Wang, ``Separation of corn seeds images based on
  threshold changed gradually,'' \emph{Transactions of the Chinese Society of
  Agricultural Engineering}, vol.~27, no.~7, pp. 200--204, 2011.

\bibitem{miller2017plantjournal}
N.~D. Miller, N.~J. Haase, J.~Lee, S.~M. Kaeppler, N.~de~Leon, and E.~P.
  Spalding, ``A robust, high-throughput method for computing maize ear, cob,
  and kernel attributes automatically from images,'' \emph{The Plant Journal},
  vol.~89, no.~1, pp. 169 -- 178, 2017.

\bibitem{grift2017biosystems}
T.~E. Grift, W.~Zhao, M.~A. Momin, Y.~Zhang, and M.~O. Bohn, ``Semi-automated,
  machine vision based maize kernel counting on the ear,'' \emph{Elsevier
  Biosystems Engineering}, vol. 164, pp. 171--180, 2017.

\bibitem{khaki2021deepcorn}
S.~Khaki, H.~Pham, Y.~Han, A.~Kuhl, W.~Kent, and L.~Wang, ``Deepcorn: A
  semi-supervised deep learning method for high-throughput image-based corn
  kernel counting and yield estimation,'' \emph{Elsevier Knowledge-Based
  Systems}, vol. 218, p. 106874, 2021.

\bibitem{10.5555/992362.2845017}
A.~M. Reza, ``Realization of the contrast limited adaptive histogram
  equalization (clahe) for real-time image enhancement,'' \emph{J. VLSI Signal
  Process. Syst.}, vol.~38, no.~1, p. 35–44, Aug. 2004.

\bibitem{resnet}
K.~He, X.~Zhang, S.~Ren, and J.~Sun, ``Deep residual learning for image
  recognition,'' in \emph{IEEE Conference on Computer Vision and Pattern
  Recognition}, 2015, pp. 770--778.

\bibitem{alvares2013koppen}
C.~Alvares, J.~Stape, P.~Sentelhas, J.~Gon{\c{c}}alves, and G.~Sparovek,
  ``K{\"o}ppen’s climate classification map for brazil,'' \emph{Stuttgart
  Meteorologische Zeitschrift}, vol.~22, no.~6, pp. 711--728, 2013.

\bibitem{adam}
D.~Kingma and J.~Ba, ``Adam: A method for stochastic optimization,''
  \emph{International Conference on Learning Representations}, 2014.

\bibitem{mann1947test}
H.~B. Mann and D.~R. Whitney, ``On a test of whether one of two random
  variables is stochastically larger than the other,'' \emph{JSTOR - The Annals
  of Mathematical Statistics}, pp. 50--60, 1947.

\bibitem{vargha1998kruskalhtest}
A.~Vargha and H.~D. Delaney, ``The kruskal-wallis test and stochastic
  homogeneity,'' \emph{Journal of Educational and Behavioral Statistics},
  vol.~23, no.~2, pp. 170--192, 1998.

\bibitem{simonyan2013deep}
K.~Simonyan, A.~Vedaldi, and A.~Zisserman, ``Deep inside convolutional
  networks: Visualising image classification models and saliency maps,''
  \emph{arXiv preprint arXiv:1312.6034}, 2013.

\end{thebibliography}

\end{sloppypar}
\end{document}